\pdfoutput=1

\documentclass[11pt]{article}

\usepackage[preprint]{acl}

\usepackage{times}
\usepackage{latexsym}

\usepackage[T1]{fontenc}

\usepackage[utf8]{inputenc}

\usepackage{microtype}

\usepackage{inconsolata}

\usepackage{graphicx}

\usepackage{hyperref}       %
\usepackage{url}            %
\usepackage{booktabs}       %
\usepackage{amsfonts}       %
\usepackage{nicefrac}       %
\usepackage{xcolor}         %
\usepackage{algorithm}      %
\usepackage{algpseudocode}  %
\usepackage{amsmath}        %
\usepackage{amssymb}        %
\usepackage{enumitem} %
\usepackage{multirow}
\usepackage{graphicx}   %
\usepackage{multirow}   %
\usepackage{pifont}
\usepackage{array}
\usepackage{tabularx}
\usepackage{longtable}
\usepackage{ltablex}
\usepackage{setspace}
\keepXColumns
\usepackage[most]{tcolorbox}

\newcommand{\cmark}{\ding{51}}%
\newcommand{\xmark}{\ding{55}}%

\definecolor{prompt}{RGB}{223, 223, 192}
\definecolor{prompt-frame}{RGB}{137, 137, 90}
\definecolor{prompt2}{RGB}{223, 223, 192}
\definecolor{prompt2-frame}{RGB}{137, 137, 90}
\definecolor{prompt3}{RGB}{212, 238, 179}
\definecolor{prompt3-frame}{RGB}{117, 146, 77}
\definecolor{prompt4}{RGB}{212, 238, 179}
\definecolor{prompt4-frame}{RGB}{117, 146, 77}
\definecolor{prompt5}{RGB}{212, 238, 179}
\definecolor{prompt5-frame}{RGB}{117, 146, 77}
\definecolor{whitesmoke}{RGB}{245, 245, 245}  %

\tcbset{
    prompt_reasoning/.style={
        enhanced,
        breakable,
        colback=prompt,        %
        colframe=prompt-frame,            %
        fontupper=\small,
        boxrule=1pt,                       %
        arc=4mm,                           %
        left=0.5mm, right=0.5mm, top=0.5mm, bottom=0.5mm, %
        boxsep=2pt,                        %
        before skip=8pt, after skip=8pt, %
        overlay={
            \node[anchor=north west, xshift=4pt, text=white] at (frame.north west) {};
        },
        title={\textbf{Prompt used for $M_R$'s generation of reasoning chains for SFT data creation.}},
        coltitle=white,
        fonttitle=\bfseries\footnotesize
    }
}
\tcbset{
    prompt_injection/.style={
        enhanced,
        breakable,
        colback=prompt2,        %
        colframe=prompt2-frame,            %
        fontupper=\small,
        boxrule=1pt,                       %
        arc=4mm,                           %
        left=0.5mm, right=0.5mm, top=0.5mm, bottom=0.5mm, %
        boxsep=2pt,                        %
        before skip=8pt, after skip=8pt, %
        overlay={
            \node[anchor=north west, xshift=4pt, text=white] at (frame.north west) {};
        },
        title={\textbf{Prompt used for $M_I$'s evidence integration for SFT data creation.}},
        coltitle=white,
        fonttitle=\bfseries\footnotesize
    }
}

\tcbset{
    prompt_model/.style={
        enhanced,
        breakable,
        colback=prompt2,        %
        colframe=prompt2-frame,            %
        fontupper=\small,
        boxrule=1pt,                       %
        arc=4mm,                           %
        left=0.5mm, right=0.5mm, top=0.5mm, bottom=0.5mm, %
        boxsep=2pt,                        %
        before skip=8pt, after skip=8pt, %
        overlay={
            \node[anchor=north west, xshift=4pt, text=white] at (frame.north west) {};
        },
        title={\textbf{System prompt for \ourmethod{}. The actual system prompt for each model prepends the corresponding model's original system prompt before this prompt.}},
        coltitle=white,
        fonttitle=\bfseries\footnotesize
    }
}

\usepackage{amsmath,amsfonts,bm}

\def\eqref#1{equation~\ref{#1}}

\def\1{\bm{1}}

\DeclareMathAlphabet{\mathsfit}{\encodingdefault}{\sfdefault}{m}{sl}
\SetMathAlphabet{\mathsfit}{bold}{\encodingdefault}{\sfdefault}{bx}{n}

\newcommand{\ourmethod}{\texttt{CARE}}

\title{Improving Context Fidelity via Native Retrieval-Augmented Reasoning}

\author{
    \textbf{Suyuchen Wang}\textsuperscript{1,3}\thanks{Equal contribution.},
    \textbf{Jinlin Wang}\textsuperscript{2}\footnotemark[1],
    \textbf{Xinyu Wang}\textsuperscript{4}\footnotemark[1],
    \textbf{Shiqi Li}\textsuperscript{2},
    \textbf{Xiangru Tang}\textsuperscript{5},\\
    \textbf{Sirui Hong}\textsuperscript{2},
    \textbf{Xiao-Wen Chang}\textsuperscript{4},
    \textbf{Chenglin Wu}\textsuperscript{2}\thanks{Corresponding authors. Contact via email: bang.liu@umontreal.ca, alexanderwu@deepwisdom.ai.},
    \textbf{Bang Liu}\textsuperscript{1,3,6}\footnotemark[2]
    \\[0.5em]
    \textsuperscript{1}DIRO, Universit\'e de Montr\'eal \quad
    \textsuperscript{2}MetaGPT \quad
    \textsuperscript{3}Mila - Quebec AI Institute \quad \\
    \textsuperscript{4}McGill University \quad
    \textsuperscript{5}Yale University \quad
    \textsuperscript{6}Canada CIFAR AI Chair
}

\begin{document}
\maketitle
\begin{abstract}

Large language models (LLMs) often struggle with context fidelity, producing inconsistent answers when responding to questions based on provided information. Existing approaches either rely on expensive supervised fine-tuning to generate evidence post-answer or train models to perform web searches without necessarily improving utilization of the given context. We propose \textbf{\ourmethod{}}, a novel native retrieval-augmented reasoning framework that teaches LLMs to explicitly integrate in-context evidence within their reasoning process with the model's own retrieval capabilities. Our method requires limited labeled evidence data while significantly enhancing both retrieval accuracy and answer generation performance through strategically retrieved in-context tokens in the reasoning chain. Extensive experiments on multiple real-world and counterfactual QA benchmarks demonstrate that our approach substantially outperforms supervised fine-tuning, traditional retrieval-augmented generation methods, and external retrieval solutions. This work represents a fundamental advancement in making LLMs more accurate, reliable, and efficient for knowledge-intensive tasks.\footnote{Homepage: \url{https://foundationagents.github.io/CARE}.}

\end{abstract}

\section{Introduction}

\begin{figure}[h!]
    \centering
    \includegraphics[width=0.95\linewidth]{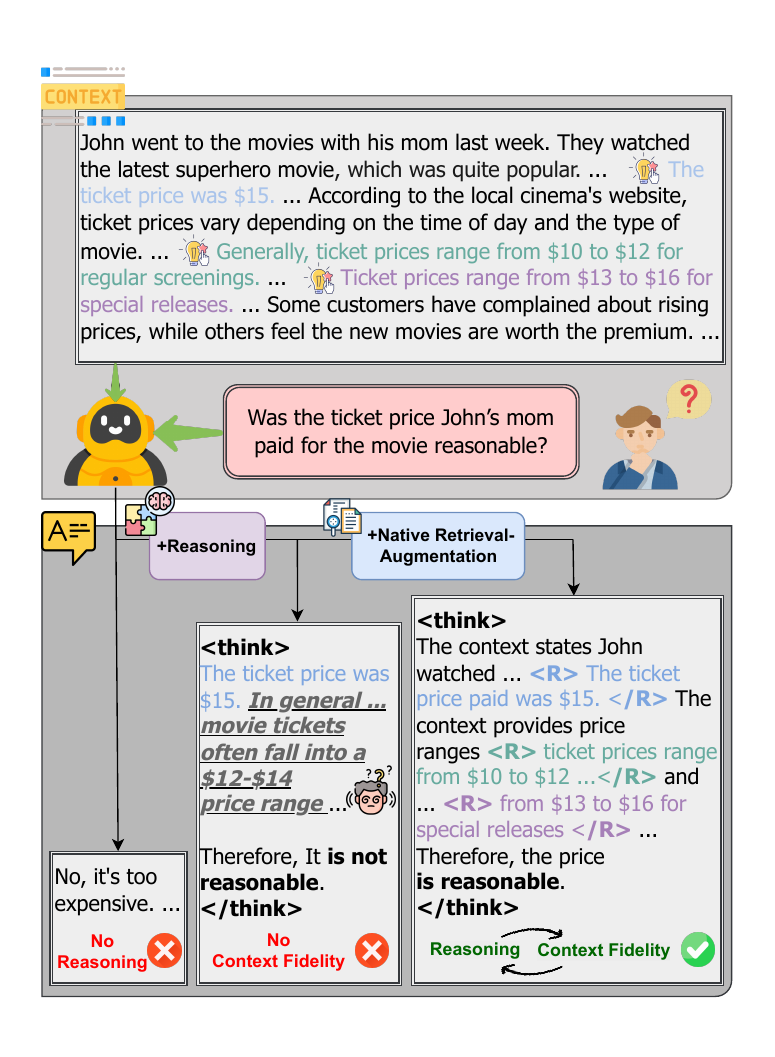}
    \caption{The Comparison among direct generation, reasoning-based generation, and reasoning with integrated in-context facts.}
    \label{fig:overview}
\end{figure}

Large language models (LLMs) have demonstrated impressive performance in a wide range of natural language tasks~\citep{minaee2024large, liu2025advances}, yet continue to struggle with a fundamental challenge: maintaining fidelity to the context provided when answering questions~\citep{talukdar2024improving}. This \textit{context hallucination problem}~\citep{chang2024booookscore, hu2024refchecker0, liu-etal-2025-towards} is particularly pronounced in knowledge-intensive tasks where precise information retrieval and accurate reasoning are paramount. When LLMs generate answers that contradict or fabricate information relative to the input context, user trust declines, and the practical utility of these systems decreases considerably.

Current approaches to addressing this challenge fall into two broad categories, each with significant limitations. The first category employs retrieval-augmented generation (RAG) for evidence retrieval~\citep{DBLP:journals/corr/abs-2312-10091,DBLP:conf/eacl/WangYW24}. Although this approach can improve explainability, it usually requires extensive labeled datasets with ground-truth evidence spans, making it prohibitively expensive to scale across diverse domains and languages. In addition, the extra retriever module and the vector database create excessiveness to the model architecture.

The second category leverages external retrieval mechanisms, allowing models to search for relevant information beyond their parametric knowledge~\citep{hsu2024grounding,nguyen2024reinforcement}. Although effective in accessing up-to-date or specialized information, these approaches frequently underutilize the rich context already provided by users, which often contains the most relevant information for their specific scenarios. Furthermore, external retrieval introduces additional latency, complexity, and potential inconsistencies between the retrieved content and the original context.

In this paper, we introduce a fundamentally different approach: \textbf{native retrieval-augmented reasoning}. Rather than treating retrieval and reasoning as separate processes, our method teaches LLMs to dynamically identify and incorporate relevant evidence from the input context directly within their reasoning chain. This approach leverages the inherent \textit{native} language understanding capabilities of LLMs to perform in-context retrieval without additional indexing or embedding systems, while simultaneously enhancing the reasoning process through explicit evidence integration.

Based on the aforementioned approach, we introduce the \textbf{Context-Aware Retrieval-Enhanced reasoning (\ourmethod{})} framework. The \ourmethod{} framework requires limited labeled evidence data and operates through a two-phase training process: an initial supervised fine-tuning (SFT) phase that establishes the evidence integration pattern, followed by a reinforcement learning (RL) phase that refines the self-retrieval mechanism through retrieval-aware rewards. Crucially, we implement a curriculum learning strategy that enables the model to progressively adapt from simple to complex reasoning tasks, extending beyond the initial training distribution without requiring additional labeled data.

Our main contributions are as follows.

\begin{itemize}
    \item We introduce \textbf{native retrieval-augmented reasoning}, a novel paradigm that organically combines in-context retrieval with structured reasoning to improve context fidelity and reduce hallucinations.
    \item We present a curated dataset for training models to perform evidence-integrated reasoning, which we have open sourced to facilitate further research in this area.
    \item We propose \textbf{\ourmethod{}}, a comprehensive implementation that combines native retrieval-augmented reasoning with curriculum learning to handle diverse question-answering scenarios without additional labeled data.
    \item Through extensive experiments across multiple real-world and counterfactual QA benchmarks, we demonstrate that our approach substantially outperforms vanilla SFT, traditional RAG methods, and comparable models lacking in-context retrieval mechanisms in both evidence retrieval and answer accuracy.
\end{itemize}

Our work represents a significant advancement in making LLMs more accurate, reliable, and efficient for knowledge-intensive tasks, particularly when relevant information is already present in the input context. By teaching models to explicitly retrieve and reason with contextual evidence, we establish a stronger foundation for context-faithful language generation.

\section{Related Work}

\subsection{LLM Reasoning on Question-Answering Tasks}

Large language models (LLMs) have demonstrated impressive capabilities in complex reasoning tasks \cite{wei2022chain,cobbe2021training,ouyang2022training}. Recent work has explored various prompting strategies to improve reasoning, including chain of thought prompting \cite{wei2022chain}, which guides models to generate intermediate reasoning steps before producing final answers, and its variants such as zero-shot-CoT \cite{kojima2022large} and self-consistency \cite{wang2022self}. More structured approaches include tree-of-thought \cite{yao2023tree}, graph-of-thought \cite{besta2024graph}, ReAct \cite{yao2023react}, and least-to-most prompting \cite{zhou2022least}. Despite these advances, LLMs still struggle to maintain context coherence when reasoning about long or noisy inputs \cite{xu2023retrieval,li2024alr,fei2024retrieval}.

\subsection{Retrieval-Augmented Generation}

Traditional retrieval-augmented generation (RAG) methods \cite{DBLP:conf/icml/GuuLTPC20,DBLP:conf/nips/LewisPPPKGKLYR020} enhance LLM by retrieving relevant passages from external corpora, alleviating the limitations of fixed parametric memory. This framework has been widely adopted for knowledge-intensive tasks \cite{DBLP:conf/acl/Xiong0LZ24,wang2025finsagemultiaspectragfinancial}. Recent work has improved retrieval quality through techniques such as query expansion \cite{DBLP:conf/emnlp/WangYW23}, re-ranking \cite{DBLP:conf/acl/VuI0CWWTSZLL24}, and filtering \cite{DBLP:conf/iclr/AsaiWWSH24}, while others focus on robustness to noisy retrievals \cite{DBLP:conf/iclr/YoranWRB24}. In-context retrieval methods aim to reuse relevant spans from the input sequence itself \cite{DBLP:journals/corr/abs-2312-10091,DBLP:conf/eacl/WangYW24}. However, both external and in-context RAG fundamentally rely on indexing and embedding-based retrieval pipelines, limiting their adaptability to complex or evolving contexts.

\subsection{RL-Enhanced LLM Retrieval}

Reinforcement learning (RL) has emerged as a powerful paradigm for optimizing LLM retrieval strategies \cite{humphreys2022large,tu2024reinforcement,hsu2024grounding}. Unlike traditional retrieval methods, RL-based approaches can learn adaptive retrieval policies that optimize for task-specific rewards \cite{kulkarni2024reinforcement,zhuang2025rank,jin2025search}.
Recent work has explored the use of RL to train retrieval policies that maximize answer accuracy \cite{hsu2024grounding,nguyen2024reinforcement}, combining the strengths of parametric knowledge and non-parametric retrieval \cite{mallen2022not,humphreys2022large,farahani2024deciphering}.
Several approaches have used feedback mechanisms to improve retrieval quality, including relevance feedback \cite{zhou2023enhancing} and iterative refinement \cite{chen2024learning}. However, most existing approaches still maintain a separation between the retrieval mechanism and the core reasoning process, potentially limiting the model's ability to integrate retrieved information in a context-aware manner.

\section{The \ourmethod{} Method}

\begin{figure*}[ht!]
    \centering
    \includegraphics[width=1.0\textwidth]{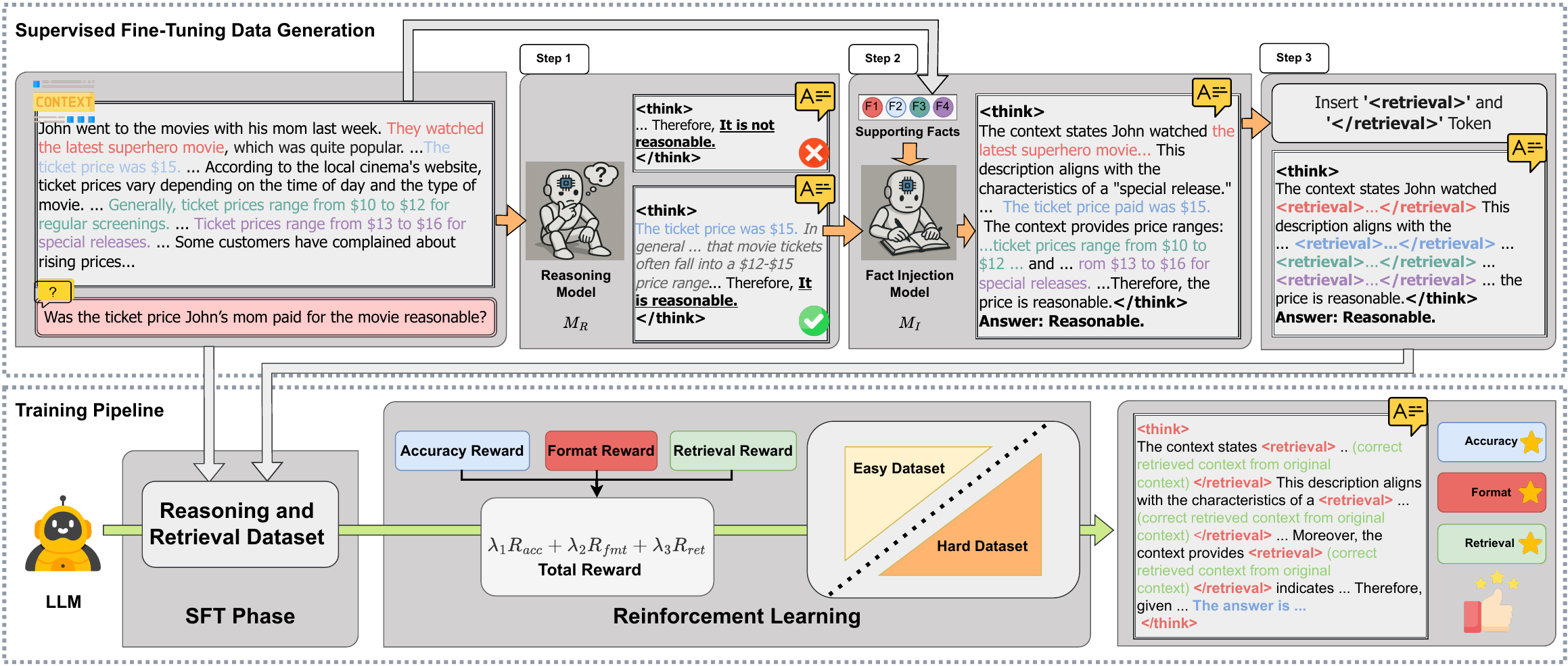}
    \caption{Illustration of the training data creation and two-phase training process of \ourmethod{}. The upper part depicts the SFT data generation pipeline including fact injection and special tokens insertion within the reasoning content. The lower part shows the SFT training process and the reinforcement learning (RL) training with multiple rewards.}
    \label{fig:pipeline}
\end{figure*}

\subsection{Overview}
We present the \ourmethod{} Method, a reasoning framework that enables LLMs to autonomously conduct native retrieval from the input context without relying on any external retrieval modules or tools, and integrate evidence retrieved by LLM's native capabilities into the reasoning process. By allowing the model to perform native retrieval, \ourmethod{} better utilizes the language understanding capability of LLMs and user input while reducing the reliance on potentially expensive tool calling, while introducing native retrieval results in the reasoning can both improve the model's context loyalty and improve the reasoning process by utilizing curated evidence. Figure ~\ref{fig:overview} illustrates the comparison between direct generation, reasoning-based inference, and \ourmethod{}.

Labeling supporting facts for QA datasets is expensive, and thus \ourmethod{}'s design aims at reducing the reliance on such labels. Thus, the framework is designed to consist of two training phases: a supervised fine-tuning (SFT) phase followed by a reinforcement learning (RL) phase.

The first SFT phase is designed to improve the efficiency of RL training by familiarizing the model with the target output format with retrieved facts in the reasoning process. In this phase, the model is fine-tuned on a self-curated dataset comprising reasoning chain enriched with golden in-context retrieval snippets, guiding the model to align reasoning with context-derived evidence (Section \ref{subsec:sft_phase}).

The RL phase refines the self-retrieval mechanism through retrieval-aware rewards, reinforcing evidence consistency and logical coherence with the help of native retrieval results from the input context throughout the reasoning process (Section \ref{subsec:rl_phase}). This phase further develops the model’s ability to identify and integrate supporting facts within the context, ensuring alignment across multi-hop reasoning steps using only question-answer pairs without golden supporting facts.

Together, these two phases form a structured framework that integrates retrieval within reasoning, improving context loyalty, retrieval accuracy, and answer correctness in QA tasks.

\subsection{Problem Formulation}

We formally define our target problem as $(Q, C) \rightarrow A$, where $Q$ is a user query, $C$ is a long context containing sparse information relevant to answering $Q$, and $A$ is the generated answer. This formulation specifically targets scenarios where the identification of key information within lengthy contexts and the utilization of sparse key information are the primary bottlenecks.

\subsection{The Supervised Fine-Tuning Phase}
\label{subsec:sft_phase}
The SFT phase establishes evidence integration by injecting retrieval tokens within structured reasoning steps. This phase uses an existing QA dataset with labeled supporting facts to ease the ``cold-start'' problem of the RL training phase, familiarizing the model with the targeted output format, the native retrieval process, and the chain-of-thought reasoning with retrieved facts as support.

Specifically, based on a given QA dataset with context and supporting facts from the context, we introduce a pipeline to generate reasoning chains interleaved with evidence.
The data generation pipeline operates sequentially through three stages: reasoning step generation, evidence integration, and retrieval token insertion, which is illustrated in the upper part of Figure~\ref{fig:pipeline}.

The data generation pipeline processes the input $(C, Q)$, where $C$ is the context and $Q$ is the query, through three stages:

\paragraph{Reasoning Step Generation.}  

The SFT dataset generation is based on an existing training dataset $\mathcal{D}_{\text{original}} = \{(Q_i, C_i, A_i, S_i)\}_{i=1}^{N_{\text{original}}}$, where the $i$-th instance contains a query $Q_i$, a context $C_i$, a ground truth answer $A_i$ and a series of labeled supporting facts from $C_i$: $S_i = \{s_i^1, s_i^2, ..., s_i^{m_i}\}$. For each instance, a reasoning model $M_R$ generates an initial reasoning response $R_{i,A}$ based on $(C_i, Q_i)$:
\begin{equation}
R_{i,A} = M_R(C_i, Q_i)
\end{equation}

For each output, we examine whether the generated answer matches the ground truth, and only responses that correctly answer the query $Q$ are retained to ensure logical consistency.  
However, some correct responses might be derived from the internal knowledge of the model rather than from the input context. This disloyalty to the context potentially increases the risk of hallucinations. To better align the reasoning process with the input context, while $R_A$ establishes a structured reasoning format, it may not align with the evidence of the context, necessitating further integration of the evidence in the next stage. For the $i$-th selected instance, we extract the reasoning process within the \textsc{<think>} \textsc{</think>} tokens to form the reasoning chain $N_i$.

\paragraph{Evidence Integration.}

To ensure that the reasoning aligns with the input context and to mitigate potential hallucinations from $R_A$, this stage integrates supporting facts $S_i$ into the reasoning chain $N_i$.
A fact injection model $M_I$ then refines the initial reasoning $R_A$ by incorporating these specific facts $F$. The output of the evidence integration process is conditioned on the query $Q_i$, the initial reasoning chain $N_i$, and the ground truth supporting facts $S_i$:
\begin{equation}
R_{i,I} = M_I(Q_i, N_i, S_i)
\end{equation}
In this formulation, $M_I$ focuses on weaving the supporting facts $S_i$ into the existing reasoning structure of $N_i$ to produce $R_{i,I}$. This step explicitly grounds the reasoning process in the supplied evidence, reducing the reliance on the model's internal knowledge when abundant context is given. After generation, we only keep the instances where $R_{i,I}$ contains all the supporting facts provided.
For the $i$-th kept instance, we retrieve the output reasoning chain with evidence integration as $E_i$, which is more robustly supported by factual statements in the context than $N_i$.

\paragraph{Retrieval Token Insertion.}

Lastly, a pair of structural marker tokens is introduced to explicitly denote the supporting fact spans in the reasoning chain to assist the further training process. The newly added tokens \textsc{<retrieval>} \textsc{</retrieval>} is inserted around key evidence segments in $R_I$, forming the final structured response $E_I^*$.

Ultimately, we obtain our SFT training set $\mathcal{D}_{\text{SFT}} = \{(Q_i, C_i, A_i, E_i^*)\}_{i=1}^{N_{\text{SFT}}}$, establishing a consistent format for the subsequent RL phase.

\subsection{Reinforcement Learning Phase} \label{subsec:rl_phase}

The reinforcement learning phase refines the self-retrieval mechanism established in the SFT phase by aligning the model’s outputs with contextual evidence through Group Relative Policy Optimization (GRPO) \cite{DBLP:journals/corr/abs-2402-03300}. This phase leverages a curriculum learning strategy to gradually transition the model from basic to advanced reasoning tasks while applying retrieval-aware rewards to promote evidence consistency and logical coherence. 

The detailed implementation of the reinforcement learning phase, including curriculum adjustment, reward computation, and policy updates, is described in Algorithm~\ref{alg:curriculum_grpo}.

\paragraph{The GRPO Algorithm.}  
GRPO optimizes the policy by evaluating multiple sampled outputs at the group level rather than individual actions. Given a query $q$, a set of outputs $\{ o_1, \ldots, o_G \}$ is sampled from the old policy $\pi_{\theta_{\text{old}}}$. The objective function is defined as:
{\small
\begin{align}
\label{eqn:grpo}
J_{\text{GRPO}}(\theta) = \ & \mathbb{E}_{q\sim\mathcal{D}, \{ o_i \}_{i=1}^G \sim \pi_{\theta_{\text{old}}(\cdot|q)}} \Big[\Big[
\frac{1}{G} \sum_{i=1}^G \frac{1}{|o_i|} \sum_{t=1}^{|o_i|} \nonumber \\
& \min \left[ w_{i,t} \hat{A}_{i,t}, \text{clip}\left( r_{i,t}, 1 - \epsilon, 1 + \epsilon \right) \hat{A}_{i,t} \right]
 \nonumber \\
& - \beta D_{\text{KL}}\left(\pi_{\theta} \parallel \pi_{\text{ref}}\right)\Big]
\end{align}}
where the importance ratio \( w_{i,t} \) is defined as:
\begin{equation}
    w_{i,t} = \frac{\pi_{\theta}(o_{i,t} | q, o_{i,<t})}{\pi_{\theta_{\text{old}}}(o_{i,t} | q, o_{i,<t})}
\end{equation}
and the advantage function $\hat{A}_{i,t}$ is defined as:
\begin{equation}
\hat{A}_{i,t} = \frac{r(q, o_i) - \text{mean}\left(\{r(q, o_i)\}_{i=1}^G\right)}{\text{std}\left(\{r(q, o_i)\}_{i=1}^G\right)}
\end{equation}

The clip function constrains the importance ratio to be within $[1-\epsilon, 1+\epsilon]$ to avoid overconfidence. The KL divergence \( D_{\text{KL}}(\pi_{\theta} \parallel \pi_{\text{ref}}) \) serves as a regularization term, preventing excessive divergence from the reference policy. Group-level evaluation in GRPO effectively promotes evidence alignment across multiple outputs, reinforcing the retrieval consistency within the reasoning process.

\paragraph{Reward Design.}
To encourage the model to retrieve relevant information from the context and dynamically integrate them into the reasoning chain, we propose the retrieval reward for \ourmethod{} training. More specifically, the retrieval reward encourages the model when it outputs the expected \textsc{<retrieval>} \textsc{</retrieval>} pair, and all text within these pairs exists in the context. Although a rather loose constraint, the retrieval reward allows the model to make better use of the context in reasoning without ground-truth retrieval data.

Furthermore, since we introduce a new pair of tokens in the reasoning process, we slightly changed the format reward proposed in~\citet{deepseek-ai2025deepseekr1}, which now pushes the model to reason with pairs \textsc{<think>} \textsc{</think>} and \textsc{<retrieval>} \textsc{</retrieval>}. Similarly, the accuracy reward quantifies the correctness of the generated response by calculating the token F1 score between the extracted generated answer and the ground truth answer for the QA tasks.

In general, the reward function in the RL phase is formulated as a weighted sum of three components, each aimed at a distinct aspect of the retrieval and alignment of the reasoning.
\begin{equation}
    R_{\text{total}} = \lambda_1 R_{\text{acc}} + \lambda_2 R_{\text{fmt}} + \lambda_3 R_{\text{ret}}
    \label{eqn:rewards}
\end{equation}

The weighting coefficients \( \lambda_1, \lambda_2, \lambda_3 \) control the relative emphasis on factual accuracy, structural consistency, and context fidelity.

\begin{algorithm}[t!]
\small  %
\caption{Curriculum RL with \ourmethod{} Rewards.}
\label{alg:curriculum_grpo}
\algrenewcommand\algorithmicindent{1em}  %
\begin{algorithmic}[1]
\Require Datasets $\mathcal{D}_{\text{easy}}$, $\mathcal{D}_{\text{hard}}$, policy $\pi_{\theta}$, reference policy $\pi_{\text{ref}}$, clip range $\epsilon$, KL coefficient $\beta$, initial ratio $\alpha = 1.0$, total steps $T$
\Ensure Updated policy parameters $\theta$
\For{each training step $t$}
    \State Sample query $q$ with probability $\alpha$ from $\mathcal{D}_{\text{easy}}$ and $1 - \alpha$ from $\mathcal{D}_{\text{hard}}$
    \State Sample outputs $\{ o_i \}_{i=1}^G$ from $\pi_{\theta_{\text{old}}}(q)$
    \For{each output $o_i$}
        \State Extract retrieval spans $S$ from $o_i$
        \State Compute rewards with Eq.~\ref{eqn:rewards}
        \For{each token $t$ in $o_i$}
            \State Compute importance ratio $r_{i,t} = \frac{\pi_{\theta}(o_{i,t})}{\pi_{\theta_{\text{old}}}(o_{i,t})}$
            \State Update objective with Eq.~\ref{eqn:grpo}
        \EndFor
    \EndFor
    \State Apply KL penalty: \\
    \ \ \ \ \ \ \ \ $J_{\text{GRPO}} \gets J_{\text{GRPO}} - \beta \sum_{t} \pi_{\theta}(o_t) \log \frac{\pi_{\theta}(o_t)}{\pi_{\text{\text{ref}}}(o_t)}$
    \State Update parameters $\theta \gets \theta + \eta \nabla_{\theta} J_{\text{GRPO}}$
    \State Adjust curriculum ratio $\alpha \gets \max(0, 1 - \eta t / T)$
\EndFor
\State \Return $\theta$
\end{algorithmic}
\end{algorithm}

\begin{table*}[!t]
\centering
\resizebox{0.85\textwidth}{!}{
\begin{tabular}{c|l|cccc|c}
\toprule
\multicolumn{1}{c}{Model} &\multicolumn{1}{c}{Method} & MFQA & HotpotQA & 2WikiMQA & MuSiQue & Average\\
\midrule
\multirow{5}{*}{LLaMA-3.1 8B} & Original & \underline{45.57} & \underline{54.64} & 45.87 & 32.08 & 44.54 \\
 & ReSearch & / & / & / & / & / \\
 & R1-Searcher & 28.44 & 53.71 & \underline{67.10} & \underline{41.41} & \underline{47.67} \\
 & CRAG & 44.04 & 37.88 & 25.95 & 24.10 & 32.99 \\
 & \ourmethod{} & \textbf{49.94} & \textbf{63.09} & \textbf{75.29} & \textbf{51.00} & \textbf{59.83} \\
 \midrule
\multirow{5}{*}{Qwen2.5 7B} & Original & 46.94 & \underline{58.47} & 46.96 & 30.78 & 45.79\\
 & ReSearch & 32.45 & 54.24 & 55.78 & \textbf{47.61} & 47.52 \\
 & R1-Searcher & 28.36 & 55.43 & \underline{65.79} & \underline{47.09} & \underline{49.17} \\
 & CRAG & \underline{47.90} & 43.97 & 33.00 & 28.44 & 38.33 \\
 & \ourmethod{} & \textbf{48.11} & \textbf{63.45} & \textbf{70.11} & 45.57 & \textbf{56.81} \\
 \midrule
\multirow{5}{*}{Qwen2.5 14B} & Original & 47.58 & \underline{61.94} & \underline{59.05} & \underline{37.99} & \underline{51.64} \\
 & ReSearch & / & / & / & / & / \\
 & R1-Searcher & / & / & / & / & / \\
 & CRAG & \textbf{50.89} & 44.74 & 34.68 & 28.17 & 39.62 \\
 & \ourmethod{} & \underline{48.81} & \textbf{67.75} & \textbf{78.68} & \textbf{51.27} & \textbf{61.63} \\
 \bottomrule
\end{tabular}
}
\caption{Evaluation on the real-world QA datasets. The results are grouped by the base LLM used. The best and second-best results for each base model and dataset are labeled in \textbf{bold} and \underline{underline}, respectively. Slash (/) indicates that the model does not have an official checkpoint or support for this model.}
\label{tab:qa}
\end{table*}

\paragraph{Curriculum Learning Strategy.}  
QA datasets exhibit significant variation in context and answer lengths. To gradually adapt our model to diverse dataset characteristics other than the one used for SFT, we implement a curriculum learning strategy transitioning from short-context / short-answer QA to long-context / multihop long-answer QA. This structured progression mitigates catastrophic forgetting while enhancing retrieval capabilities in increasing complexity.

We train with two QA datasets: $\mathcal{D}_{\text{easy}} = \{(Q_i, C_i, A_i)\}_{i=1}^{N_{\text{easy}}}$ and $\mathcal{D}_{\text{hard}} = \{(Q_i, C_i, A_i)\}_{i=1}^{N_{\text{hard}}}$, where $\mathcal{D}_{\text{hard}}$ contains longer contexts, longer answers, and requires more complex reasoning than $\mathcal{D}_{\text{easy}}$. Training begins exclusively with $\mathcal{D}_{\text{easy}}$, then gradually incorporates instances from $\mathcal{D}_{\text{hard}}$.

At each training step $t$, we sample instances using a Bernoulli trial with a time-varying probability. The mixing ratio $\alpha_t$ decreases linearly according to $\alpha_t = \max(0, 1 - \eta \cdot \frac{t}{T})$, 
where $\eta$ is a scaling factor that controls the speed of transition. The sampling probabilities are $p_{\text{easy}} = \alpha_t$ and $p_{\text{hard}} = 1 - \alpha_t$, ensuring that the model maintains short-context retrieval capabilities while learning to aggregate evidence in multiple paragraphs.

\section{Experiment Settings}

We evaluate our proposed \ourmethod{} method through comprehensive experiments across multiple LLM families and sizes in two distinct QA categories: real-world long-context QA and counterfactual multihop QA.

\subsection{Datasets, Benchmarks and Metrics}
\label{sec:datasets}

\paragraph{Training Datasets.}
We generate the SFT data mentioned in Section~\ref{subsec:sft_phase} based on the HotpotQA training set~\citep{yang2018hotpotqa} owing to its annotations of supporting facts. During SFT data generation, DeepSeek-R1~\citep{deepseek-ai2025deepseekr1} and DeepSeek-V3~\citep{deepseek-ai2024deepseek0v3} are used as the reasoning model $M_R$ and the fact injection model $M_I$, respectively. The resulting SFT dataset contains 7,739 instances with the retrieval-augmented reasoning chain labeled. For RL training, we select DROP~\citep{dua2019drop} as $\mathcal{D}_{\text{easy}}$ and MS MARCO~\citep{nguyen2016msmarco} as $\mathcal{D}_{\text{hard}}$.

\paragraph{Evaluation Datasets.}
We evaluate in-context retrieval accuracy and whether learned retrieval-augmented reasoning improves answer quality using single-passage and multi-passage datasets from LongBench~\citep{bai-etal-2024-longbench}, including \textbf{MultiFieldQA-En}~\citep{bai-etal-2024-longbench}, \textbf{HotpotQA}~\citep{yang2018hotpotqa}, \textbf{2WikiMQA}~\citep{ho-etal-2020-constructing}, and \textbf{MuSiQue}~\citep{trivedi-etal-2022-musique}. Following LongBench's protocol, we report F1 scores for all datasets.

Furthermore, to evaluate context fidelity when presented with information contradicting the model's parametric knowledge, we utilize \textbf{CofCA}~\citep{wu2025cofca}, a benchmark containing modified counterfactual Wikipedia snippets. This directly tests whether our native retrieval-augmented reasoning improves adherence to provided context regardless of pre-trained biases. We report F1 performance consistent with the original CofCA evaluation metrics.

\begin{table}[!t]
\centering
\footnotesize
\resizebox{0.83\columnwidth}{!}{
\begin{tabular}{c|l|c}
\toprule
\multicolumn{1}{c}{Model} &\multicolumn{1}{c}{Method} & CofCA\\
\midrule
\multirow{3}{*}{LLaMA-3.1 8B} & Original & \underline{48.14}\\
 & R1-Searcher & 45.25 \\
 & \ourmethod{} & \textbf{61.83} \\
 \midrule
\multirow{5}{*}{Qwen2.5 7B} & Original & \underline{58.38} \\
 & ReSearch &  47.32 \\
 & R1-Searcher & 43.61 \\
 & CRAG &  56.01 \\
 & \ourmethod{} & \textbf{64.56} \\
 \midrule
\multirow{3}{*}{Qwen2.5 14B} & Original & \underline{64.40} \\
 & CRAG & 51.99  \\
 & \ourmethod{} & \textbf{67.75}  \\
 \bottomrule
\end{tabular}
}
\caption{Evaluation on the counterfactual QA task. The results are grouped by the base LLM used. The best and second-best results for each base model and dataset are labeled in \textbf{bold} and \underline{underline}, respectively.
}
\label{tab:counterfactual}
\end{table}

\subsection{Models and Baselines}

We compare \ourmethod{} with a series of learned reasoning strategies and RAG methods based on three commonly used public LLMs: Qwen-2.5 Instruct 7B and 14B~\citep{qwen2024qwen205}, and LLaMA-3.1 8B~\citep{grattafiori2024llama}, which covers different model families and sizes.

\paragraph{Original Model.} For each dataset, we test the performance of the original LLM with their corresponding default system prompt and chat template.

\paragraph{RL-Based Online Retrieval.} Existing dynamic retrieval approaches typically leverage reinforcement learning to train models to autonomously conduct web searches rather than directly extract from the provided context. We compare our method against two recent RL-based online search methods: ReSearch~\citep{chen2025research} and R1-Searcher~\citep{song2025r1}, both of which enable models to strategically access external knowledge during reasoning. Note that in our model selection, ReSearch only provides a checkpoint for Qwen2.5 7B, and R1-Searcher only provides a checkpoint for LLaMA-3.1 8B and Qwen2.5 7B.

\paragraph{RAG Methods.} We also compare with CRAG~\citep{yan2024corrective}, a corrective RAG method that uses a lightweight evaluator to improve in-context retrieval with online search. Note that in our model selection, CRAG only provides a checkpoint for Qwen2.5 7B and 14B.

\subsection{Experiment Settings}
\label{sec:experiment_settings}
For all three models, the SFT training process follows LLaMA-Factory~\citep{zheng2024llamafactory}'s default LoRA SFT setting\footnote{\url{https://github.com/hiyouga/LLaMA-Factory/blob/main/examples/train_lora/llama3_lora_sft.yaml}}, and the RL training process follows verl~\citep{sheng2024hybridflow0}'s default GRPO setting\footnote{\url{https://github.com/volcengine/verl/blob/main/examples/grpo_trainer/run_qwen2-7b.sh}}. For the Curriculum RL training, we use the hyperparameters $\lambda_1=0.7$, $\lambda_2=0.1$, $\lambda_3=0.2$, and $\eta=1$. All experiments are done with either 8$\times$A800-SXM4-80GB or 8$\times$H100 80GB. Detailed experiment settings are included in Appendix~\ref{appx:exp_details}.

\section{Results and Analysis}

\subsection{Question-Answering Performance}
\label{sec:real_world_qa}

Table~\ref{tab:qa} shows that \ourmethod{} consistently outperforms baselines in all model sizes. With LLaMA-3.1 8B, our method achieves +15.29\% average F1 improvement over the original model, with the strongest gains in multi-hop tasks (2WikiMQA +29.42\%, MuSiQue +18.92\%). Similar patterns appear with the Qwen2.5 models. Even when not highest (Qwen2.5 7B on MuSiQue and Qwen2.5 14B on MFQA), \ourmethod{} remain competitive with the best baseline. These results demonstrate that native retrieval-augmented reasoning significantly enhances performance by effectively integrating in-context evidence during reasoning, especially for complex multi-hop questions. We also include a token cost analysis in Appendix~\ref{appx:token_cost}. While \ourmethod{} generates longer outputs due to the reasoning chains, which is a common characteristic of GRPO-trained models, it eliminates the overhead of external API calls and database retrievals required by baseline methods.

\subsection{Counterfactual QA Performance}

\begin{table*}[!t]
\centering
\resizebox{0.98\textwidth}{!}{
\begin{tabular}{l|cccc|ccccc|c}
\toprule
Settings & SFT & RL & Ret. & Cur. & MFQA & HotpotQA & 2WikiMQA & MuSiQue & CofCA & Average \\
\midrule
Baseline & \xmark & \xmark & \xmark & \xmark & \underline{46.64} & 58.47 & 46.96 & 30.78 & 58.38 & 48.25 \\
SFT Only & \cmark & \xmark & \xmark & \xmark & 42.24 & 47.08 & 61.51 & 33.82 & 59.21 & 48.77 \\
No Ret. & \cmark & \cmark & \xmark & \xmark & 37.66 & 62.59 & \underline{70.57} & 43.85 & 57.26 & 54.39 \\
No Cur. & \cmark & \cmark & \cmark & \xmark & 38.33 & \textbf{64.10} & \textbf{70.69} & \textbf{47.49} & \underline{60.60} & \underline{56.24} \\
\ourmethod{} & \cmark & \cmark & \cmark & \cmark & \textbf{48.11} & \underline{63.45} & 70.11 & \underline{45.57} & \textbf{64.56} & \textbf{58.36} \\
\bottomrule
\end{tabular}
}
\caption{Ablation studies on the QA tasks based on Qwen2.5 7B. The best and second-best results for each base model and dataset are labeled in \textbf{bold} and \underline{underline}, respectively. ``Ret.'' stands for retrieval reward in Equation~\ref{eqn:rewards}, and ``Cur.'' stands for curriculum learning in Algorithm~\ref{alg:curriculum_grpo}.}
\label{tab:ablation}
\end{table*}

\begin{figure*}[t!]
    \centering
    \includegraphics[width=\textwidth]{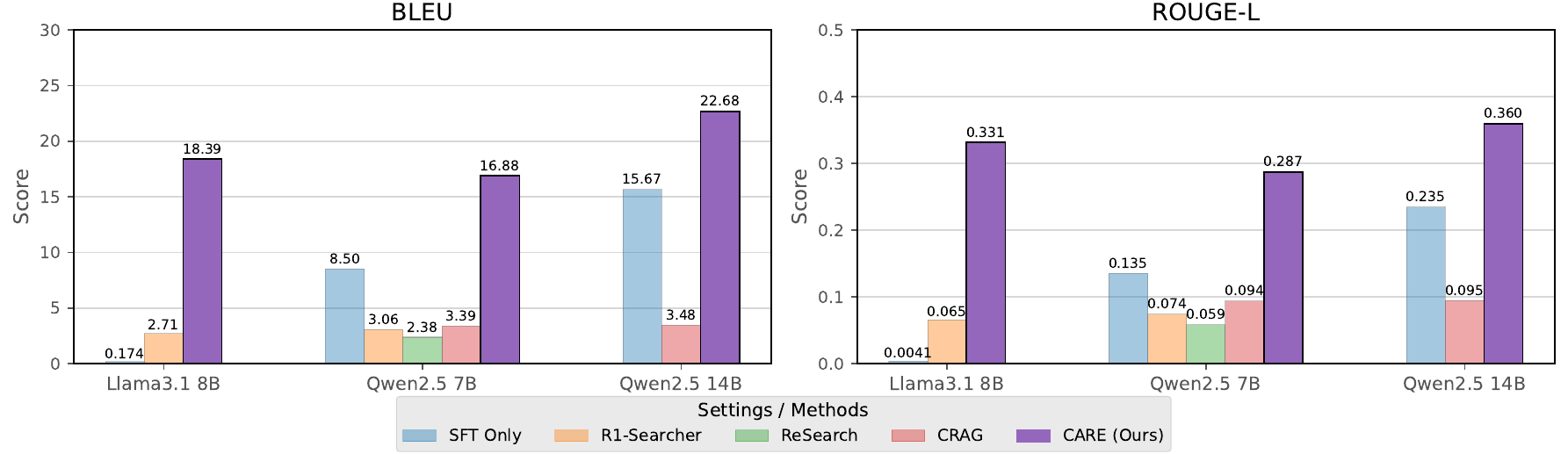} %
    \caption{Comparison of model performance across different settings for BLEU and ROUGE-L metrics. Our proposed methods, \ourmethod{}, demonstrate improved scores.}
    \label{fig:retrieval}
\end{figure*}

In Table~\ref{tab:counterfactual}, we report the results on the CofCA counterfactural QA task. \ourmethod{} consistently delivers the strongest performance, with significant gains on LLaMA-3.1 8B (+13.69\%). In particular, traditional online search methods underperform compared to original models in this task, suggesting that external retrieval can be counterproductive when context contradicts parametric knowledge. \ourmethod{} demonstrates superior context fidelity by explicitly integrating natively extracted in-context evidence in the reasoning process, and can make even greater gains compared to the baselines when encountering unseen information in the context.

\subsection{Ablation Studies}

We provide results based on Qwen2.5 7B in Table~\ref{tab:ablation}. In this table, we include three additional settings: \textbf{(1) SFT only}, where the model is only trained with the first SFT phase without RL training; \textbf{(2) No retrieval reward}, where after the SFT phase, the model undergoes GRPO training with the same reasoning-encouraging reward used in DeepSeek-R1~\citep{deepseek-ai2025deepseekr1}; and \textbf{(3) No curriculum learning}, where the RL training phase uses only $\mathcal{D}_{\text{easy}}$.

SFT alone only offers marginal benefits, while adding RL training substantially improves performance, highlighting the importance of reinforcement learning for QA reasoning. Both methods with native in-context reasoning (``No Cur.'' and \ourmethod{}) consistently outperform the vanilla R1-like GRPO approach (``No Ret.''), showing that retrieval-augmented reasoning enhances performance by grounding reasoning in contextual evidence. While ``No Cur.'' performs well on multi-hop datasets, curriculum learning provides better balance across diverse types of QA, particularly improving performance on long-form answering (MFQA) and counterfactual scenarios (CofCA). This shows that curriculum learning successfully adapts the model to various types of question while maintaining strong performance on complex reasoning tasks, all without requiring additional labeled data beyond the initial SFT phase.

\subsection{Evidence Retrieval Evaluation}

In this section, we evaluate \ourmethod{}'s ability to accurately retrieve and incorporate supporting evidence for question-answering. Due to the lack of ground-truth supporting fact annotations in standard QA datasets, we focus our evaluation on the LongBench HotpotQA benchmark. For this analysis, we align each instance in LongBench's HotpotQA test set with its corresponding entry in the original HotpotQA dataset, using the original supporting fact annotations as ground truth for evaluation. We report SacreBLEU~\citep{post-2018-call} and ROUGE-L F1~\citep{lin-2004-rouge}. Figure~\ref{fig:retrieval} presents our comparative results in different model configurations. In all settings, \ourmethod{} consistently achieves the highest BLEU and ROUGE-L scores. We observe that performance scales with model size across all methods, with Qwen2.5 14B showing the strongest results. However, the relative improvement from \ourmethod{} remains consistent regardless of the scale and family of the model, suggesting that our approach effectively enhances the context fidelity regardless of the underlying model architecture.

\section{Conclusion}

We introduce \ourmethod{}, a native retrieval-augmented reasoning framework that improves context fidelity in LLM by teaching models to dynamically identify and integrate evidence within their reasoning process. This approach improves how LLMs interact with context, while requiring limited labeled evidence. Experiments on multiple general and counterfactual QA benchmarks demonstrated that \ourmethod{} consistently outperforms existing approaches, including the vanilla SFT method and traditional RAG methods in both answer generation and evidence extraction. This work represents an important step toward more reliable AI systems that make better use of available context without requiring expensive retrieval infrastructure.

\section*{Limitations}

Although \ourmethod{} shows significant improvements in context fidelity and question answering performance, several important limitations should be acknowledged. First, the native retrieval-augmented reasoning mechanism, while effective for in-context information, cannot access external knowledge beyond the provided context. For scenarios requiring information not present in the input, our approach would need to be combined with external retrieval systems like RAG, potentially complicating the overall architecture.

Second, while we evaluate comprehensively across multiple QA benchmarks, the evaluation primarily focuses on multi-hop general-domain reasoning questions. The effectiveness of \ourmethod{} for more abstract reasoning, numerical computation, creative generation tasks, or domain-specific tasks remains to be thoroughly investigated.

Finally, although our method improves context fidelity, it does not completely eliminate the possibility of hallucinations, especially when the input contains ambiguous or contradictory information. Future work should address these challenges while expanding the approach to a broader range of language understanding and generation tasks.

\section*{Ethical Considerations}

Our research improves context fidelity in language models, potentially reducing hallucinations in critical applications like education and healthcare. The supervised fine-tuning dataset we developed, built upon HotpotQA (which follows the CC BY-SA 4.0 license), will be shared under the same CC BY-SA 4.0 license to promote transparency and reproducibility. Although our method improves fidelity to the provided context, it cannot guarantee complete factual accuracy, especially when the input itself contains inaccuracies or contradictions. We acknowledge both the environmental impact of the computational resources used in training and the possibility that models may inherit biases from training data despite improved context fidelity. Researchers who implement our approach should perform appropriate fairness evaluations before deployment in sensitive applications.

\section*{Acknowledgement}

This research was enabled in part by the support provided by Calcul Qu\'ebec\footnote{\url{https://www.calculquebec.ca/}.} and the Digital Research Alliance of Canada\footnote{\url{https://alliancecan.ca/}.}.

\bibliography{custom}

\newpage
\appendix

\section{Token Cost Efficiency Analysis}
\label{appx:token_cost}

In this section, we provide a detailed token cost analysis for the real-world QA experiments (Section~\ref{sec:real_world_qa}) in Table~\ref{tab:token_analysis}. While \ourmethod{} generates longer outputs due to the reasoning chains, which is a common characteristic of GRPO-trained models, it eliminates the overhead of external API calls and database retrievals required by baseline methods.

\begin{table}[h!]
\centering
\small

\resizebox{0.5\textwidth}{!}{
\begin{tabular}{l|l|cccc}
\toprule
Model & Method & MFQA & HotpotQA & 2Wiki & MuSiQue \\
\midrule
\multirow{3}{*}{LLaMA 8B} & Original & 19.5 & 8.5 & 7.5 & 7.4 \\
& R1-S. & 296+2012 & 278+2058 & 293+2125 & 313+2436 \\
& \ourmethod{} & 564 & 656 & 608 & 848 \\
\midrule
\multirow{4}{*}{Qwen 7B} & Original & 24.8 & 6.0 & 6.1 & 9.3 \\
& CRAG & 24+212 & 7+411 & 9+201 & 8+470 \\
& ReSearch & 276+2054 & 275+2271 & 308+2814 & 290+2492 \\
& \ourmethod{} & 566 & 633 & 560 & 942 \\
\bottomrule
\end{tabular}
}
\caption{Average output tokens per query on each real-world QA dataset. Numbers in format $x$+$y$ indicate model output + retrieval overhead. R1-S. stands for R1-Searcher.}
\label{tab:token_analysis}
\end{table}

\section{Experiment Details}
\label{appx:exp_details}

\subsection{Implementation Details}

All models are implemented based on pre-trained checkpoints provided by the Huggingface Transformers library~\citep{wolf2019huggingface0s}. We use LLaMA-Factory~\citep{zheng2024llamafactory} for the SFT phase. In this phase, we train each model on our curated SFT dataset for 3 epochs with the AdamW optimizer~\citep{loshchilov2018decoupled}. The training progress adopts a warmup cosine scheduler with a maximum learning rate 0.0001 and a warmup ratio of 0.1. The effective batch size is 64. LoRA~\citep{hu2022lora} is applied with $r=8$ and $\alpha=16$. The training process uses the ZeRO-2 optimizer~\citep{rajbhandari2019zero0}. For the RL phase, we adopt the verl framework~\citep{sheng2024hybridflow0} for GRPO training. We used a training batch size of 1024. The Adam optimizer was employed with a learning rate of 1e-6. For policy optimization, we use GRPO as the advantage estimator and incorporated KL divergence regularization with a coefficient of 0.001 using the low-variance KL estimator. We set the mini-batch size to 256. The model was trained for 350 steps with 5 response samples per prompt. For distributed training, we deployed Fully Sharded Data Parallel (FSDP)~\citep{DBLP:journals/pvldb/ZhaoGVLHXWSOSDB23} across 8 GPUs on a single node with tensor parallelism of size 2. All experiments are done with either 8$\times$A800-SXM4-80GB or 8$\times$H100 80GB.

\section{System Prompts}

We provide the system prompts used in the dataset creation process and the \ourmethod{} in the following.

\begin{tcolorbox}[prompt_reasoning]
You're an expert reader. Your goal is to read a context to answer a question. Note that during your thinking process, before you make *any reasoning step that requires retrieving information from the context*, summarize what information you would need to complete this reasoning step, such as "I need to know X for this" or similar phrases before you reason about the context. This will help you to be more systematic in your reasoning process. Put your final answer as a minimum phrase or word at the end after ``Answer:''.
    
    Context:
    \{context\}
    
    Question:
    \{question\}
\end{tcolorbox}

\begin{tcolorbox}[prompt_injection]
I'll provide you with a question, a reasoning process to solve this question, and several evidence sentences. Insert *all* evidence sentences into the reasoning process at appropriate locations and give me the updated reasoning process. Each evidence sentence usually should be placed just before any conclusions or deductions that depend on it. The evidence sentences may need to be distributed throughout different parts of the reasoning and may appear more than once. *Do not modify any evidence sentences* - insert them exactly as provided. Return only the completed reasoning process without explanations or additional text scaffolds.
    
    Question:
    \{question\}
    
    Reasoning process:
    \{reasoning\_content\}
    
    Evidence sentences (One sentence per line):
    \{evidence\_sentence\_string\}
    
    The rewritten reasoning process:
\end{tcolorbox}

\begin{tcolorbox}[prompt_model]
You FIRST think about the reasoning process as an internal monologue and then provide the final answer. The reasoning process MUST BE enclosed within <think> </think> tags. WITHIN the thinking process, make reference to the relevant texts in the prompt that provide critical information to move the reasoning process forward. The referenced texts MUST BE enclosed within <retrieval> </retrieval> tags, and MUST BE placed within the reasoning process only. The final answer MUST BE put at the end of the response after ``Answer:''.
\end{tcolorbox}

\section{Case Study}

We select two samples from LongBench's HotpotQA dataset and provide the complete question, answer, supporting facts, prediction, and whether the generated answer is correct.

\begin{table*}[!t]
\centering
\resizebox{\textwidth}{!}{

{\small
\begin{tabular}{m{1.2cm}|m{1cm}|m{3.2cm}|m{9.3cm}|m{0.8cm}}
\toprule
Question & Answer & Supporting Facts & Prediction & Correct \\
\midrule
The song that came out ahead of The Wanted's ``Lightning'' on the UK Singles Chart was featured on which album? & 
Talk That Talk &
[\texttt{"}``We Found Love'' is a song recorded by Barbadian singer Rihanna from her sixth studio album, ``Talk That Talk'' (2011).\texttt{"}, \texttt{"}It was written and produced by, and features Scottish DJ Calvin Harris; the song was also featured on his third album ``18 Months'' (2012).\texttt{"}, \texttt{"}``We Found Love'' premiered on September 22, 2011, on the Capital FM radio station in the United Kingdom, and was made available to download on the same day as the lead single from ``Talk That Talk''.\texttt{"}, \texttt{"}``We Found Love'' is an uptempo electro house song, with elements of techno and Europop.\texttt{"}, \texttt{"}The song's lyrics speak of a couple who ``found love in a hopeless place''.\texttt{"}] & 
<think>\texttt{\textbackslash n}Okay, let's try to figure out the answer to this question. The question is asking which album featured the song that came out ahead of The Wanted's ``Lightning'' on the UK Singles Chart.\texttt{\textbackslash n}\texttt{\textbackslash n}First, I need to find out the details about The Wanted's ``Lightning'' from the passages provided. Looking at the passage for The Wanted's discography, it mentions that their third single from the second album ``Battleground'' was ``Lightning,'' released in October 2011. <retrieval>The song debuted at number two on the UK Singles Chart, behind Rihanna's ``We Found Love''.</retrieval> So, that means the song that was number one before ``Lightning'' was ``We Found Love'' by Rihanna.\texttt{\textbackslash n}\texttt{\textbackslash n}Now, the question is about the album that features ``We Found Love.'' I need to check where ``We Found Love'' is mentioned. In the passages, there's a section about Rihanna's ``We Found Love'' from her album ``Talk That Talk (2011.'' The album is listed as her sixth studio album. The song itself is part of that album. <retrieval>``We Found Love'' is a song by Barbadian singer Rihanna from her sixth studio album, Talk That Talk (2011.</retrieval> So the answer should be ``Talk That Talk.''\texttt{\textbackslash n}\texttt{\textbackslash n}Wait, but let me confirm. The relevant information from the third passage (The Rihanna section for ``We Found Love'' clearly states that it's from ``Talk That Talk (2011),'' and that's her sixth studio album. The song's debut was with Rihanna's album. Therefore, the answer is ``Talk That Talk.''\texttt{\textbackslash n}</think>\texttt{\textbackslash n}\texttt{\textbackslash n}Answer: Talk That Talk &
Yes  \\
\midrule
Yau Ma Tei North is a district of a city with how many citizens? & 
7.2 million &
[\texttt{"}Hong Kong, officially the Hong Kong Special Administrative Region of the People's Republic of China, is an autonomous territory south to Mainland China and east to Macao in East Asia.\texttt{"}, \texttt{"} With around 7.2 million Hong Kongers of various nationalities in a territory of 1,104 km, Hong Kong is the world's fourth most densely populated country or territory.\texttt{"}] & 
<think>\texttt{\textbackslash n}Okay, let's tackle this question. The question is asking how many citizens Yau Ma Tei North is a district of. First, I need to figure out which city Yau Ma Tei North is a district of. From the context, the term ``Yau Ma Tei North'' is mentioned in several passages. Let me check.\texttt{\textbackslash n}\texttt{\textbackslash n}Looking at the passages, the one titled ``Yau Ma Tei North (constituency)'' states that Yau Ma Tei North is one of the 19 constituencies in the Yau Tsim Mong District of Hong Kong. \texttt{\textbackslash n}<retrieval>Yau Ma Tei North is one of the 19 constituencies in the Yau Tsim Mong District of Hong Kong which was first created in 1982 and recreated in 2015.</retrieval>\texttt{\textbackslash n}That's in the context of Hong Kong's administrative divisions. The next step would be to find out the population of Hong Kong. \texttt{\textbackslash n}\texttt{\textbackslash n}I recall that there's a detailed section in the ``Hong Kong'' passages about the population. Let me check that part. The Hong Kong entry mentions that the Census and Statistics Department estimated Hong Kong's population at 7,413,070 in 2021.\texttt{\textbackslash n}<retrieval>The Census and Statistics Department estimated Hong Kong's population at 7,413,070 in 2021. The overwhelming majority (91.6\%) is Han Chinese, most of whom are Taishanese, Teochew, Hakka, and other Cantonese peoples.</retrieval>\texttt{\textbackslash n}\texttt{\textbackslash n}Wait, but the question is about Yau Ma Tei North, which is a district within Hong Kong. The population of Hong Kong as a whole is 7.4 million, but the key point here is whether Yau Ma Tei North itself has a population figure. The Yau Ma Tei North constituency's entry says it covers Yau Ma Tei and has an estimated population of 12,817. \texttt{\textbackslash n}<retrieval>The constituency loosely covers Yau Ma Tei with the estimated population of 12,817.</retrieval>\texttt{\textbackslash n}So that's 12,817 people. But the question is phrased as ``Yau Ma Tei North is a district of a city with how many citizens?'' So the target is the parent city of Yau Ma Tei North. Since Yau Ma Tei North is a district in Hong Kong, Hong Kong is the city. Therefore, the population of Hong Kong is the answer, which is 7,413,070 as per the latest data provided.\texttt{\textbackslash n}</think>\texttt{\textbackslash n}\texttt{\textbackslash n}To determine the population of the city (Hong Kong) containing Yau Ma Tei North, I need to know the administrative level of Yau Ma Tei North. The ``Yau Ma Tei North (constituency)'' section states it belongs to Yau Tsim Mong District. Next, the ``Hong Kong'' passages mention the district is part of the Yau Tsim Mong District and its population. The main Hong Kong entry (under ``Demographics'') gives the 2021 population as 7,413,070. However, Yau Ma Tei North is a specific constituency (the 19th out of 19) within Yau Tsim Mong District, with an estimated population of 12,817. The question likely refers to the broader city of Hong Kong's population, not the smaller constituency.  \texttt{\textbackslash n}Answer: 7413070 & 
No  \\
\bottomrule
\end{tabular}
}
}
\caption{Case study examples from LongBench's HotpotQA test set showing \ourmethod{}'s reasoning process. The table presents the original questions, ground-truth answers and supporting facts, along with the complete model predictions including the reasoning chains within \texttt{<think>} tags and integrated retrieval spans within \texttt{<retrieval>} tags. The "Correct" column indicates whether the model's final answer matches the ground truth using F1 score evaluation.}
\end{table*}

\end{document}